\documentclass[a4paper,12pt]{article}
\usepackage[utf8]{inputenc}
\usepackage[T2A]{fontenc}
\usepackage{graphicx} 
\usepackage{amsmath,amssymb} 
\usepackage{tikz}
\usepackage{url}
\usepackage{cite}
\usepackage{caption}
\usepackage{pgfplots}
\usepackage{multirow}
\usepackage{float}
\usepackage{setspace}
\usepackage{booktabs}
\usepackage{todonotes}
\usepackage[a4paper,
            bindingoffset=0.0in,
            left=1.2in,
            right=1.2in,
            top=1in,
            bottom=1in,
            footskip=.5in]{geometry}
\usetikzlibrary{pgfplots.groupplots,positioning,decorations.pathmorphing}

\usepackage{array}
\newcolumntype{H}{>{\setbox0=\hbox\bgroup}c<{\egroup}@{}}

\usepackage{cmap}
\DeclareSymbolFont{cyrillic}{T2A}{cmr}{m}{}
\SetSymbolFont{cyrillic}{bold}{T2A}{cmr}{bx}{}
\DeclareMathSymbol{\kyrgyzy}{\mathord}{cyrillic}{179}
\DeclareMathSymbol{\kyrgyzo}{\mathord}{cyrillic}{176}
\DeclareMathSymbol{\kyrgyzn}{\mathord}{cyrillic}{173}
\DeclareMathSymbol{\kyrgyzY}{\mathord}{cyrillic}{147}
\DeclareMathSymbol{\kyrgyzO}{\mathord}{cyrillic}{144}
\DeclareMathSymbol{\kyrgyzN}{\mathord}{cyrillic}{141}

\title{Syntactic Transfer to Kyrgyz \\ Using the~Treebank Translation Method}

\author{
Anton Alekseev$^{1,2,3,4}$, Alina Tillabaeva$^5$, \\
Gulnara Dzh. Kabaeva$^4$, Sergey I. Nikolenko$^{6,1}$~\thanks{The work was supported by a grant from the Russian Science Foundation \#22-11-00135, ``Research and development of technologies for processing and analyzing multimodal unstructured data from various sources and their applicability for solving economic and social problems''.}\\[10pt]
$^1$ \small Steklov Mathematical Institute at St.~Petersburg, St. Petersburg, Russia\\
$^2$ \small St. Petersburg University, St. Petersburg, Russia\\
$^3$ \small Kazan (Volga Region) Federal University, Kazan, RT, Russia\\
$^4$ \small Kyrgyz State Technical University n.~a. I.~Razzakov, Bishkek, Kyrgyzstan\\
$^5$ \small Independent Researcher, Bishkek, Kyrgyzstan\\
$^6$ \small ITMO University, St. Petersburg, Russia
}

\date{}

\begin{document}

\maketitle

\begin{abstract}
The Kyrgyz language, as a low-resource language, requires significant effort to create high-quality syntactic corpora. This study proposes an approach to simplify the development process of a syntactic corpus for Kyrgyz.
We present a tool for transferring syntactic annotations from Turkish to Kyrgyz based on a treebank translation method. The effectiveness of the proposed tool was evaluated using the TueCL treebank. The results demonstrate that this approach achieves higher syntactic annotation accuracy compared to a monolingual model trained on the Kyrgyz KTMU treebank.
Additionally, the study introduces a method for assessing the complexity of manual annotation for the resulting syntactic trees, contributing to further optimization of the annotation process.
\end{abstract}

\section{Introduction}

Kyrgyz language, like many other low-resource languages (LRLs)~\cite{mirzakhalov2021evaluating,veitsman2024recentadvancementschallengesturkic,alekseev2024kyrgyznlpchallengesprogressfuture}, has recently garnered increasing attention from research communities aiming to enhance machine-readable resources and tools for the related studies. One of the critical challenges in this domain is the development of syntactic corpora (treebanks), which demands significant efforts from experts and typically requires a large span of time.

In this work, we propose a semi-automatic approach to syntactic analysis as a potential solution to this problem. We perform cross-lingual syntactic parsing from a source language to Kyrgyz leveraging treebank translation method. By utilizing preliminary annotations generated by models, linguists can focus on refining and correcting these annotations, thus accelerating the overall annotation process. We believe that the approach has the potential to substantially reduce the effort required to produce high-quality syntactic corpora of sufficient size.

We demonstrate the feasibility of this approach and present a specialized tool for syntactic transfer of the annotations from a resource-rich source language (Turkish) to the target language (Kyrgyz). The shared grammatical features of the selected languages, particularly their word order and agglutinative nature, facilitate cross-linguistic syntactic transfer.

Finally, we evaluate the proposed system using a recently published Kyrgyz treebank annotated within the Universal Dependencies (UD) framework. 
We have prepared a Python package implementing our method: \url{https://github.com/alexeyev/tratreetra}. The package allows to plug in any dependency parser, morphological analyzer, or word aligner.

\section{Related Work}\label{sec:relatedwork}

\subsection{Kyrgyz Language in Universal Dependencies}\label{ssec:related_work_kyrgyz_ud}

Significant progress has been made in recent years in adapting the approach of the \emph{Universal Dependencies} (UD) framework~\cite{ud2021} for the Kyrgyz language. Efforts are ongoing to expand syntactically annotated corpora for Kyrgyz and to formalize labeling guidelines that address challenges such as copula tokenization, the annotation of modal words and null-headed clauses, and distinguishing between inflection and derivation. These challenges are similar to those encountered in developing UD resources for other Turkic languages and have been the subject of several recent studies~\cite{dzhumalieva2023,musazhanova2023,kasieva2023problems}. The resulting resources and guidelines will play a crucial role in training parsers and overall advancing the tools for Kyrgyz syntactic analysis.

As of November 2024, two significant Kyrgyz syntactic corpora have been developed under the UD framework:

\textbf{UD\_Kyrgyz-KTMU Treebank.} This treebank~\cite{benli2023}, in its initial version included in UD, contained $781$ sentences. By late October 2024, it had grown to $2'480$ sentences annotated within the dependency grammar framework~\cite{tesniere1959elements}. The dataset, hereafter referred to as \textit{KTMU}, is described in~\cite{benli2024}.

\textbf{Kyrgyz-TueCL Treebank.} This treebank contains $145$ sentences, including $20$ from the \textit{Cairo} dataset~\cite{nivre2015towards} and approximately $100$ provided by the \emph{UD Turkic Group}\footnote{Additional information can be found at \url{https://github.com/ud-turkic}.}. Each sentence is accompanied by translations into English, Turkish, and Azerbaijani, making it a part of the broader UD Turkic Treebank initiative~\cite{unification2024,washington2024strategies}.

These resources represent crucial steps toward creating robust syntactic tools for the Kyrgyz language and integrating it into the broader Universal Dependencies ecosystem. However, the annotation approaches employed in these treebanks vary significantly. Some of these differences have been outlined in~\cite{kasieva2023problems}, particularly in the annotation of copulas (e.g., the forms <<эле>>, <<болгон>>) and modal words (e.g., <<да>>, <<эле>>, <<керек>>, <<бар>>, <<жок>>).

For instance, in the \textit{TueCL} dataset, the discourse particle <<–бы>> is treated as a separate token (e.g., <<жата-бы>>, <<б-екен>>), while in~\textit{KTMU}, this interrogative particle is not annotated separately. 

Additionally, the negative word <<эмес>> is analyzed in \textit{TueCL} as an adverb (\texttt{ADV}) functioning as an adverbial modifier (\texttt{advmod}). In contrast, in \textit{KTMU}, this word is annotated as a \texttt{VERB}. The syntactic role of the particle is inconsistently annotated, appearing with labels such as \texttt{compound:svc}, \texttt{acl}, \texttt{ccomp}, and \texttt{nmod}.

Postpositions (e.g., <<$\kyrgyzy$ч$\kyrgyzy$н>>, <<чейин>>, <<со$\kyrgyzn$>>) are also annotated differently. In \textit{TueCL}, these words are analyzed as \texttt{ADP} with an \texttt{case} relation, whereas in \textit{KTMU}, <<$\kyrgyzy$ч$\kyrgyzy$н>> and <<чейин>> are given the part-of-speech (POS) tag \texttt{ADV} and are treated as \texttt{advmod}, and <<со$\kyrgyzn$>> is annotated as \texttt{NOUN} functioning as \texttt{nmod}.

The examples presented above are not by any means an exhaustive list of annotation differences between the datasets. We present them to provide a general sense of the extensive inconsistencies present in Kyrgyz treebanks.

\subsection{Syntactic Transfer}\label{sec:syntax_transfer}

A detailed survey of existing methods for transferring syntactic parsing was presented in~\cite{review_of_synt_transfer_methods}. The authors categorize the approaches into three main types: model transfer, annotation projection, and treebank translation. The model transfer approach involves training models on source language data and then applying the trained model to parse the target language. In this case, the model is typically trained on PoS tags, sometimes enhanced with morphological features of words.

The basic version of the model transfer approach is demonstrated in~\cite{delex_model_transfer}. The authors train a model to predict syntactic roles using sentences where all words are replaced with their corresponding PoS tags. The trained model is then applied to sentences in the target language, and in the final step, the resulting parses are enriched with lexical information.

To achieve better results, lexical features can be incorporated into the model during training. This can be accomplished by adding glosses for each word~\cite{Synt_transfer_glosses,Synt_transfer_glosses_2} or utilizing multilingual vector representations~\cite{synt_trans_embeddings}.

Annotation projection is another noteworthy approach, relying on the availability of a parallel corpus for the selected languages. In this method, sentences in the source language are parsed using a monolingual parser, and the syntactic annotations are transferred to word-aligned sentences in the target language~\cite{annotation_projection}. The generated syntactic trees are then utilized to train a monolingual model for the target language. This method has already been used to parse Kyrgyz sentences from the Manas epic corpus~\cite{tillabaeva2024syntactic}.

One challenge with this approach is the syntactic differences between languages (such as between Kyrgyz and Russian in the mentioned study), which complicates the task of word alignment.

In~\cite{Multilingual_Projection_for_Parsing}, the authors demonstrate that this issue can be resolved by utilizing multiple source languages and combining multilingual projections based on weighted similarity scores between the source languages and the target language.
In~\cite{rasooli_collins_2015_density}, the authors combine annotation projection and model transfer methods. They modify sentences from the source language to match the dominant syntactic dependency patterns obtained from the annotation projection onto the target language, thus addressing syntactic distance between languages.

The third approach, known as treebank translation, is similar to annotation projection but relies on a parallel corpus generated through machine translation~\cite{treebank_translation_2014}, which can be conducted at the word or sentence level. In this paper, the authors evaluate various strategies for SMT-based translation of the source language's gold-standard treebank.
In~\cite{treebank_translation_2013}, the authors emphasize the importance of consistent annotated treebanks for cross-lingual syntactic transfer. They also create a multilingual treebank incorporating five European languages and Korean.

The treebank translation method is the closest to the goals of our study, as the currently available treebank for Kyrgyz is extremely limited in size (see Section~\ref{ssec:related_work_kyrgyz_ud}). However, our work differs from~\cite{treebank_translation_2014} in its choice of a target language from the Turkic family, for which fewer treebanks are available compared to European languages. Furthermore, efforts to unify annotation standards across languages in this group are only just beginning. Therefore, the goal of this paper is not to train a monolingual model but rather to focus on creating a tool to facilitate manual annotation.

\section{Proposed Algorithm}\label{sec:method}

The method described below represents a simple pipeline for syntactic parsing (within the dependency grammar formalism) in resource-constrained settings. It proposes leveraging models trained on related languages with richer resources. The stages of the pipeline are listed below.

\begin{enumerate}
    \item \emph{Identifying a dataset for quality evaluation:} select an appropriate dataset for assessing the target language syntactic parsing results.
    \item \emph{Selecting a model for the source language:} choose a pre-trained syntactic parsing model within the same formalism for a syntactically similar language. 
    \item \emph{Target text translation:} automatically translate the text into the source language.
    \item \emph{Annotation projection:} transfer syntactic annotations to the target language sentence using automatic word alignment (bitext alignment) and simple algorithmic transformations.
    \item \emph{Evaluation of projection:} assess the quality of the generated parse tree using standard metrics and evaluation scripts based on the original dataset. This provides an estimate of the number of corrections the annotator will need to make.
\end{enumerate}

\subsection{Selecting a Dataset for Quality Evaluation}
In Section~\ref{ssec:related_work_kyrgyz_ud}, we highlighted the most significant differences between the available Kyrgyz treebanks. In our opinion, the \emph{TueCL} treebank provides the most detailed and consistent annotations. Additionally, one of the baseline models we used was the only available Kyrgyz model, Stanza~\cite{qi2020stanza} \texttt{ktmu-nocharlm}, which was trained on the \emph{KTMU} bank. Therefore, using \emph{KTMU} for quality evaluation would not be appropriate.

\subsection{Syntactic Parsers for Turkish}\label{ssec:synt_tr}
When selecting parsers, we considered the recency and size of the training corpus, as well as the ease of use of the corresponding tool. Among the main instruments available for syntactic analysis compatible with the Universal Dependencies format and formalism, we chose parsers from the \textit{Stanza} library~\cite{qi2020stanza}, trained on treebanks from UD version 2.12: \textit{BOUN}~\cite{TurkEtAl2022} and \textit{IMST}~\cite{sulubacak2016imst,sulubacak2016universal,sulubacak2018implementing}. In our view, these datasets are closer in terms of the annotation scheme to \textit{TueCL} than others. We also tested UDPipe models trained on UD version 2.5 (the \textit{IMST} treebank).

\subsection{Translation systems}
As translation systems, we considered popular services providing machine translation: Google Translate and Yandex.Translate\footnote{Yandex.Translate had to be excluded due to numerous obviously incorrect translations (for instance, some names were replaced with unrelated nouns). It is worth noting, however, that Kyrgyz translation functionality in Yandex.Translate is still in beta.}. Additionally, we used translations generated by GPT4o~\cite{achiam2023gpt}; the corresponding prompt, which required maintaining the order and number of words, is presented in Table~\ref{tab:prompt}. While not all translations thus obtained met the requirements, the results presented in Section~\ref{sec:experiments} suggest that this method had a significant impact on the quality of the final syntactic transfer.

\begin{table}[!t]
    \centering\small
    \begin{tabular}{|p{10cm}|}\hline
\begin{verbatim}    
Here are the sentences in Kyrgyz.

Кыз досуна кат жазды.
Жамгыр жаап жатат окшойт.

<...>

Дениз уктатылды.
Алар кетти.
Ал кетти.

I want you to translate them to Turkish 
line-by-line, but you must not change 
the word order and the total number 
of words in each sentence. Do not add 
any extra comments.
\end{verbatim}\\\hline
    \end{tabular}
    \caption{A prompt to ChatGPT4o to obtain the translations of sentences in Kyrgyz from the \textit{TueCL} treebank into Turkish ``encouraging'' preserving the order and number of words in each sentence.}
    \label{tab:prompt}
\end{table}

\subsection{Syntactic Transfer}

To fill all fields adopted in the Universal Dependencies project for each word in the analysis, it is also necessary to obtain the lemma of each word. Therefore, in addition to the annotation projection described above, we implemented lemmatization using the morphological analyzer apertium-kir~\cite{washington2012finite}. For simplicity, for each word, the first analysis item provided by the tool was selected. The heuristics proposed below will also require a method for determining part-of-speech tags (high accuracy is not required yet desirable); for this purpose, apertium-kir was also used as a PoS tagger.


By this stage, the texts must be tokenized, and Turkish translations should be obtained from \textit{TueCL}, GPT-4o, and Google Translate. These translations should be enriched with Universal Dependencies (UD) syntactic annotations using the Stanza-IMST-charlm, Stanza-IMST-BERT, Stanza-BOUN-BERT models, as well as UDPipe-1 (please see Section~\ref{ssec:synt_tr}).

Next, word alignment~\cite{brown1993mathematics} should be performed between Turkish and Kyrgyz sentences, followed by transformation using heuristic algorithms to facilitate the transfer of syntactic labels and structures from the source to the target sentences. A detailed description of this process is provided below.

\subsubsection{Word Alignment}

\begin{figure}[!t]
    \centering
    \begin{tikzpicture}[every node/.style={anchor=base, font=\small}]
        \node (s1) at (0,0) {Стояли};
        \node (s2) [right=1cm of s1] {звери};
        \node (s3) [right=1cm of s2] {около};
        \node (s4) [right=1cm of s3] {двери};
        \node (s5) [right=1cm of s4] {.};
        
        \node (t1) at (-0.2,-1.0) {The};
        \node (t2) [right=.25cm of t1] {beasts};
        \node (t3) [right=.25cm of t2] {were};
        \node (t4) [right=.25cm of t3] {standing};
        \node (t5) [right=.25cm of t4] {by};
        \node (t6) [right=.25cm of t5] {the};
        \node (t7) [right=.25cm of t6] {door};
        \node (t8) [right=.25cm of t7] {.};
        
        \draw[thin] (s1) -- (t4); 
        \draw[thin] (s1) -- (t3); 
        \draw[thin] (s2) -- (t2); 
        \draw[thin] (s3) -- (t5); 
        \draw[thin] (s4) -- (t7); 
        \draw[thin] (s5) -- (t8); 
    \end{tikzpicture}    
    \caption{An example of Russian and Enlish sentences, aligned.}
\label{fig:ru_en_example}
\end{figure}
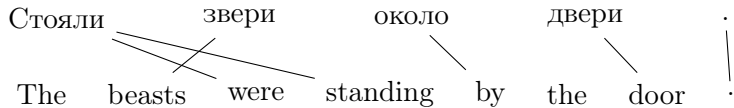

We utilized bitext alignment using SimAlign~\cite{jalili-sabet-etal-2020-simalign} to align words between Kyrgyz and Turkish in both directions.  
The result of this procedure is generally a ``many-to-many'' relation, represented as an arbitrary bipartite graph where edges connect word pairs from the two languages.
It is worth noting that a more advanced model, such as~\cite{dou2021word}, could have been used to deliver a better quality of token alignment. Additionally, the model could have been fine-tuned on sentence pairs (e.~g., by constructing a parallel Turkish-Kyrgyz corpus using machine translation). However, we opted for an interface with minimal additional configuration. For alignment, the base models employed were the multilingual RoBERTa model (XLM-R)~\cite{liu2019roberta,conneau2020unsupervised} and mBERT (a multilingual model based on BERT)~\cite{devlin2019bert}. 
An example of alignment for sentences in Russian and English is shown in Figure~\ref{fig:ru_en_example}. Note that in this example, some words (e.g., the articles ``the'') cannot be aligned to a counterpart in the other language, while other words are aligned to multiple words simultaneously (e.g., ``стояли'' and ``were standing''). Additionally, the differences in word order 
between Russian and English contribute to the ``complexity'' of the alignment. For this reason, Turkish was chosen as the source language for syntactic parsing of Kyrgyz sentences.

Thus, at this stage, we obtain an approximate alignment between sentences in Turkish and Kyrgyz, as well as dependency trees where the nodes are tokens from the Turkish sentences.

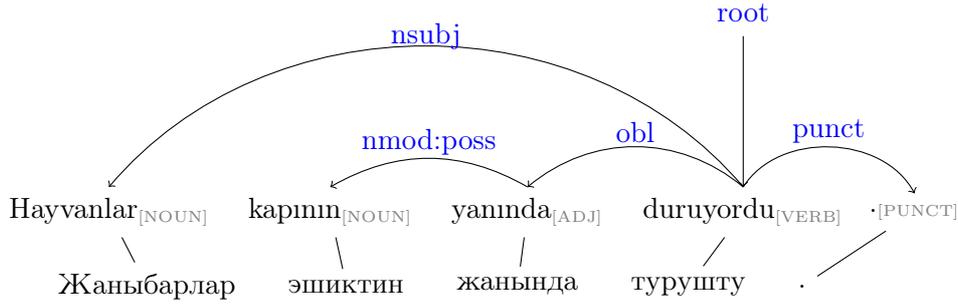
\begin{figure}[!t]
    \centering
    \begin{tikzpicture}[every node/.style={anchor=base, font=\small}]        
        \node (t1) at (0,0) { Hayvanlar\textcolor{gray}{\scriptsize$_{\mathrm{[NOUN]}}$}};
        \node (t2) [right=0.23cm of t1] {kap{\i}n{\i}n\textcolor{gray}{\scriptsize$_{\mathrm{[NOUN]}}$}};
        \node (t3) [right=0.23cm of t2] {yan{\i}nda\textcolor{gray}{\scriptsize$_{\mathrm{[ADJ]}}$}};
        \node (t4) [right=0.23cm of t3] {duruyordu\textcolor{gray}{\scriptsize$_{\mathrm{[VERB]}}$}};
        \node (t5) [right=0.05cm of t4] {.\textcolor{gray}{\scriptsize$_{\mathrm{[PUNCT]}}$}};
        
        \node (s1) at (0.5,-1.0) {Жаныбарлар};
        \node (s2) [right=0.4cm of s1] {эшиктин};
        \node (s3) [right=0.4cm of s2] {жанында};
        \node (s4) [right=0.4cm of s3] {турушту};
        \node (s5) [right=0.4cm of s4] {.};
        
        \draw[thin] (t1) -- (s1); 
        \draw[thin] (t2) -- (s2); 
        \draw[thin] (t3) -- (s3); 
        \draw[thin] (t4) -- (s4); 
        \draw[thin] (t5) -- (s5); 
        
        \draw[->, bend right=50, thin] (t4.north) to (t1.north); 
        \draw[->, bend right=40, thin] (t4.north) to (t3.north); 
        \draw[->, bend right=30, thin] (t3.north) to (t2.north); 
        \draw[->, bend left=60, thin] (t4.north) to (t5.north); 
        \draw[->, thin] ($(t4.north)+(0,2.0)$) -- (t4.north); 
                
        \node[blue] at ($(t4.north)!0.5!(t1.north)+(0,1.9)$) {nsubj};
        \node[blue] at ($(t4.north)!0.5!(t3.north)+(0,0.6)$) {obl};
        \node[blue] at ($(t3.north)!0.5!(t2.north)+(0,0.5)$) {nmod:poss};
        \node[blue] at ($(t4.north)!0.5!(t5.north)+(0,0.7)$) {punct};    
        \node[blue] at ($(t4.north)+(0,2.2)$) {root}; 
    \end{tikzpicture}
    \caption{Translations of the sentence ``Стояли звери около двери'' (en. ``The beasts were standing by the door'') into Turkish and Kyrgyz languages demonstrate a very similar word order. Above the Turkish sentence, its Universal Dependencies parse is shown~--- obtained using UDPipe~\cite{udpipe2} (model \texttt{turkish-imst-ud-2.12-230717}, trained on the UD-IMST treebank data~\cite{sulubacak2016universal,sulubacak2018implementing}).}
    \label{fig:tr_ky_example}
\end{figure}

In an ideal scenario, where the number of tokens matches and the alignment forms an identity permutation, the result would appear as shown in Figure~\ref{fig:tr_ky_example}.

However, often, this is not the case: alignments commonly include a number of edges between words at non-corresponding positions, multiple tokens aligned to a single one (as seen in Figures~\ref{fig:ru_en_example} and~\ref{fig:tr_ky_errors_example}), as well as a certain number of erroneous alignments. Therefore, we developed a set of annotation projection rules based on practical considerations, without tuning them to a specific dataset (\textit{TueCL}), as described below.

In addition to the accuracy of transferring dependency arcs and labels, structural requirements must also be taken into account and satisfied. The resulting tree must be acyclic and have a valid sentence head (or root), which is a non-trivial task. The problem arises due to Turkish root not always having a corresponding token in Kyrgyz sentences because of alignment errors or, more broadly, structural mismatches. These requirements are satisfied through the use of the following heuristics, which were implemented without tuning on the test set.

\subsubsection{Identifying the Sentence Head}

The head of the Turkish sentence is identified during the automatic parsing process, and it is checked whether this token has a corresponding token in the Kyrgyz sentence.

\begin{enumerate}
    \item If there is a single matching token for the Turkish root, the pair is excluded from the alignment, ensuring its inclusion in the final representation (the matching is constructed in the next stage; see below).
    
    \item If the Turkish root is not matched to any token, a ``greedy'' selection and assignment of the pair are performed by reverse-order traversal of the Kyrgyz sentence (considering its SOV structure), prioritizing the following in order:
    \begin{itemize}
        \item a token of the same PoS is the highest priority for the match;
        \item a verb;
        \item a noun;
        \item the first word in the sentence.
    \end{itemize}
    
    \item If the Turkish root is matched to multiple tokens, we select the one the position of which is closest to the root token's position in the Turkish sentence (the absolute value of the difference in the respective indices should be the smallest possible). This ensures that the root of the Turkish sentence is included in the final alignment.
\end{enumerate}

\subsubsection{Alignments Filtering}

For Turkish tokens with multiple alignments, PoS annotations from apertium-kir (converted to the Universal Tagset using apertium2ud~\cite{apertium2ud2023alekseev}) are utilized to filter the alignments, retaining only those that match the corresponding parts-of-speech. If this filtering procedure reduces the number of alignments to zero, the original set of alignments is preserved.

\subsubsection{Constructing the Matching}

Since the alignment can be treated as a bipartite graph, one can construct a maximum matching --- retaining edges such that no node is incident to more than one edge while maximizing the number of edges. Using the SciPy library~\cite{2020SciPy-NMeth}, the Hopcroft-Karp algorithm~\cite{hopcroft1973n} is applied to automatically build this matching, which is then used as the alignment for transferring labels. This approach eliminates the ``cycle problem'' (discussed earlier) during annotation transfer. As a result, structural correctness is ensured, though this step may potentially lead to some loss of information.

\subsubsection{Adjusting the Sentence Head (Technical Step)}

For the Kyrgyz token corresponding to the Turkish root, as determined by the constructed pairing, the \texttt{head} is manually set to $0$.

\subsubsection{Annotation Projection}

All information from the Turkish sentence, except for \texttt{id}, \texttt{Token}, and \texttt{Lemma}, is transferred to the annotation for the corresponding Kyrgyz tokens. The \texttt{head} field is then updated to reflect the structure of the dependency arcs from the syntactic parse of the Turkish sentence.

The proposed heuristic approach ensures a high degree of universality. Further possibilities for improving the annotation transfer results through heuristic tuning on a held-out dataset are discussed in Section~\ref{sec:conclusion}.

\subsection{Quality Evaluation}

Currently, there are user-friendly tools for dependency trees annotation~\cite{brat0tool,heinecke2019}, including those specifically designed for Universal Dependencies~\cite{tyers-etal:2018}. 
Standard syntactic parsing quality metrics within the framework of dependency grammar are suitable for evaluating the effort required by experts to reannotate treebanks prepared using our method.

For example, UAS (Unlabeled Attachment Score) --- a set of metrics evaluating the proportion of correctly predicted arcs (edges) --- indicates the proportion of arcs that would need to be removed (precision) and the proportion that would need to be added anew (recall). LAS (Labeled Attachment Score) is a similar but stricter metric that considers dependency relations, indicating the proportion that would need to be either re-checked or relabeled with a different type.

Similarly, the precision and recall of tokenization (creation and removal of the nodes in the tree), the precision and recall of part-of-speech tag assignments (creation and removal of UPOS tags), and other annotations can be interpreted in the same manner.

The \textit{TueCL} treebank was used for quality evaluation, as its annotation principles selected by the experts currently provide, in our opinion, the most comprehensive reflection of the syntactic features of the Kyrgyz language that need to be considered, at least for these sentences.

For reproducibility, error prevention, and overall convenience, the UD's script \texttt{tools/eval.py}~\cite{evalpy} was used as the quality evaluation tool.

    \begin{table}[!t]
    \centering\small\setlength{\tabcolsep}{4pt}
     \begin{tabular}{lllccccccccc}
        \toprule
        \multirow[c]{2}{*}{\rotatebox{90}{\textbf{MT}}} & \multirow[c]{2}{*}{\rotatebox{90}{\textbf{Alig.}}} &  & \multicolumn{3}{c}{\textbf{UAS}} & \multicolumn{3}{c}{\textbf{LAS}} & \multicolumn{3}{c}{\textbf{UPOS}} \\
        & & \textbf{Parser} &  \textbf{Pr} & \textbf{Re} & \textbf{F1} & \textbf{Pr} & \textbf{Re} & \textbf{F1} & \textbf{Pr} & \textbf{Re} & \textbf{F1} \\
        \midrule
        \multicolumn{12}{c}{Kyrgyz Language}\\\midrule
         --- & --- & St$_{\mathrm{KTMU,~nclm}}$ & 49.02 & 47.65 & 48.33 & 29.29 & 28.47 & 28.88 & 68.04 & 66.13 & \textbf{67.07}\\\midrule
         \multicolumn{12}{c}{Turkish Language}\\\midrule
\multirow[c]{8}{*}{\rotatebox{90}{GPT4o}} & \multirow[c]{4}{*}{\rotatebox{90}{SA$_{\mathrm{BERT}}$}} & St$_{\mathrm{BOUN,~BERT}}$ & 57.30 & 55.64 & 56.46 & 45.99 & 44.66 & \underline{45.31} & 61.93 & 60.14 & 61.02 \\
&  & St$_{\mathrm{IMST,~BERT}}$ & 55.04 & 53.45 & 54.23 & 43.21 & 41.96 & 42.57 & 61.83 & 60.04 & 60.92 \\
&  & St$_{\mathrm{IMST,~charlm}}$ & 51.54 & 50.05 & 50.79 & 39.61 & 38.46 & 39.03 & 61.83 & 60.04 & 60.92 \\
&  & UDPipe & 48.46 & 47.05 & 47.74 & 26.65 & 25.87 & 26.25 & 56.07 & 54.45 & 55.25 \\\cmidrule{2-12}
& \multirow[c]{4}{*}{\rotatebox{90}{SA$_{\mathrm{XLMR}}$}} & St$_{\mathrm{BOUN,~BERT}}$ & 60.39 & 58.64 & \textbf{59.50} & 48.25 & 46.85 & \textbf{47.54} & 63.17 & 61.34 & \underline{62.24} \\
&  & St$_{\mathrm{IMST,~BERT}}$ & 57.82 & 56.14 & \underline{56.97} & 45.68 & 44.36 & 45.01 & 63.07 & 61.24 & 62.14 \\
&  & St$_{\mathrm{IMST,~charlm}}$ & 54.32 & 52.75 & 53.52 & 41.87 & 40.66 & 41.26 & 63.07 & 61.24 & 62.14 \\
&  & UDPipe & 50.62 & 49.15 & 49.87 & 27.37 & 26.57 & 26.96 & 57.10 & 55.44 & 56.26 \\\midrule
\multirow[c]{8}{*}{\rotatebox{90}{Google Translate}} & \multirow[c]{4}{*}{\rotatebox{90}{SA$_{\mathrm{BERT}}$}} & St$_{\mathrm{BOUN,~BERT}}$ & 55.04 & 53.45 & 54.23 & 41.05 & 39.86 & 40.45 & 57.10 & 55.44 & 56.26 \\
&  & St$_{\mathrm{IMST,~BERT}}$ & 52.47 & 50.95 & 51.70 & 39.09 & 37.96 & 38.52 & 57.10 & 55.44 & 56.26 \\
&  & St$_{\mathrm{IMST,~charlm}}$ & 50.21 & 48.75 & 49.47 & 35.80 & 34.77 & 35.28 & 57.10 & 55.44 & 56.26 \\
&  & UDPipe & 48.25 & 46.85 & 47.54 & 23.66 & 22.98 & 23.31 & 53.81 & 52.25 & 53.02 \\\cmidrule{2-12}
& \multirow[c]{4}{*}{\rotatebox{90}{SA$_{\mathrm{XLMR}}$}} & St$_{\mathrm{BOUN,~BERT}}$ & 56.17 & 54.55 & 55.35 & 42.59 & 41.36 & 41.97 & 57.82 & 56.14 & 56.97 \\
&  & St$_{\mathrm{IMST,~BERT}}$ & 53.19 & 51.65 & 52.41 & 40.12 & 38.96 & 39.53 & 57.82 & 56.14 & 56.97 \\
&  & St$_{\mathrm{IMST,~charlm}}$ & 50.62 & 49.15 & 49.87 & 36.83 & 35.76 & 36.29 & 57.82 & 56.14 & 56.97 \\
&  & UDPipe & 47.74 & 46.35 & 47.03 & 22.84 & 22.18 & 22.50 & 54.01 & 52.45 & 53.22 \\\midrule
\multirow[c]{8}{*}{\rotatebox{90}{Transl. from \emph{TueCL}}} & \multirow[c]{4}{*}{\rotatebox{90}{SA$_{\mathrm{BERT}}$}} & St$_{\mathrm{BOUN,~BERT}}$ & 51.65 & 50.15 & 50.89 & 40.53 & 39.36 & 39.94 & 58.64 & 56.94 & 57.78 \\
&  & St$_{\mathrm{IMST,~BERT}}$ & 47.94 & 46.55 & 47.24 & 37.96 & 36.86 & 37.40 & 58.54 & 56.84 & 57.68 \\
&  & St$_{\mathrm{IMST,~charlm}}$ & 43.21 & 41.96 & 42.57 & 33.64 & 32.67 & 33.15 & 58.54 & 56.84 & 57.68 \\
&  & UDPipe & 43.42 & 42.16 & 42.78 & 22.53 & 21.88 & 22.20 & 54.42 & 52.85 & 53.62 \\\cmidrule{2-12}
& \multirow[c]{4}{*}{\rotatebox{90}{SA$_{\mathrm{XLMR}}$}} & St$_{\mathrm{BOUN,~BERT}}$ & 53.50 & 51.95 & 52.71 & 42.39 & 41.16 & 41.76 & 57.72 & 56.04 & 56.87 \\
&  & St$_{\mathrm{IMST,~BERT}}$ & 50.72 & 49.25 & 49.97 & 39.71 & 38.56 & 39.13 & 57.61 & 55.94 & 56.77 \\
&  & St$_{\mathrm{IMST,~charlm}}$ & 45.68 & 44.36 & 45.01 & 34.98 & 33.97 & 34.47 & 57.61 & 55.94 & 56.77 \\
&  & UDPipe & 45.47 & 44.16 & 44.80 & 23.66 & 22.98 & 23.31 & 53.70 & 52.15 & 52.91 \\
    \bottomrule
    \end{tabular}
    \caption{Quality ``metrics'': \textbf{Pr}~--- precision, \textbf{Re}~--- recall, \textbf{F1}~--- F-measure; SA~--- SimAlign; UDPipe~--- the UDPipe-1 parser, model \texttt{IMST-UD-2.5-191206}; St$_{[treebank],~[model]}$~--- corresponding Stanza models.}
    \label{tab:eval}
\end{table}

\section{Results}\label{sec:experiments}

All experiments were conducted on a computer with 16 GB of RAM and an Intel i7-8565U CPU @ 1.80GHz with 4 cores and 8 logical processors. The results of the experiments are presented in Tables~\ref{tab:eval} and~\ref{tab:smol_eval}.
Table~\ref{tab:smol_eval} shows metrics for which the results of all syntactic transfer models are identical. This is because these metrics evaluate parse elements obtained using apertium-kir, except for the Stanza \texttt{ktmu-noncharlm} parser, which is the only parser directly designed for Kyrgyz considered in this study. The ``Words'' metric assesses the alignment accuracy of matched words, while the ``Lemmas'' metric evaluates whether words were correctly normalized to their dictionary form.

Since the proposed text processing pipeline consists of multiple steps, it is not feasible to reliably assess the individual impact of errors at each stage on the final result. The reliability of the conclusions is also likely influenced by the small size of the test set. However, the obtained results allow us to draw the following conclusions:

\begin{itemize}
    \item for this task, the XLM-RoBERTa model demonstrates superior alignment quality in zero-shot mode compared to mBERT. For instance, in the UPOS prediction task, alignment using XLM-RoBERTa provided an advantage (up to $1.22$\%) in $8$ out of $12$ cases (except for Turkish sentences from \textit{TueCL}). Similarly, for UAS evaluations, XLM-RoBERTa outperformed in $10$ out of $12$ cases, and for LAS evaluations, in $11$ out of $12$ cases.
    
    \item for syntactic parsing tasks, models leveraging BERT-based representations consistently outperformed other tested approaches;
    
    \item determining parts of speech based on syntactic transfer was, as expected, not sufficiently effective (though it was not the primary goal of this work), hence for this task, preference should be given to apertium-kir or annotation using Stanza-KTMU-nocharlm; predicting parts-of-speech (UPOS tags) using Stanza-KTMU-nocharlm achieves a precision of $68.04$\%, recall of $66.13$\%, and an F$_1$-score of $67.07$\%; in contrast, the best results obtained through annotation projection using the Stanza-BOUN-BERT model are $\mathbf{Pr} = 63.17$\%, $\mathbf{Re} = 61.34$\%, and $\mathbf{F_1} = 62.24\%$;
    
    \item the use of machine translation with a special instruction via GPT4o provided a noticeable improvement in quality. This is clearly evident in the consistently higher F$_1$-scores for UAS, LAS, and UPOS metrics. In contrast, annotation transfer from ``original'' Turkish sentences generally performs worse, which is expected, as ``human'' translation lacks factors that could potentially and partially preserve similar word order (even at the cost of translation accuracy and grammatical correctness).
\end{itemize}

\begin{table}[!t]
    \centering
    \begin{tabular}{lp{3.8cm}cccccc}
    \toprule
     &  & \multicolumn{3}{c}{\textbf{Lemmas}} & \multicolumn{3}{c}{\textbf{Words}} \\
     & \textbf{Model} & \textbf{Pr} & \textbf{Re} & \textbf{F1} & \textbf{Pr} & \textbf{Re} & \textbf{F1} \\
    \midrule
    KY &  St$_{\mathrm{KTMU,~nclm}}$  & 74.61 & 72.53 & 73.56 & 96.81 & 94.11 & 95.44 \\
    TR &  Other (apertium-kir) & 75.82 & 73.63 & 74.71 & 97.02 & 94.21 & 95.59 \\ 
    \bottomrule
    \end{tabular}
    \caption{Quality Metrics: \textbf{Words}~--- alignment accuracy of matched words, \textbf{Lemmas}~--- how accurate the conversion of the words to their dictionary form is.}
    \label{tab:smol_eval}
\end{table}

We would like to note that our approach demonstrates superiority compared to the monolingual Kyrgyz parser Stanza-KTMU-nocharlm.
This outcome serves as an inspiration for further research and the development of more advanced parsers.

Moreover, the results obtained open up new opportunities for further improving syntax analysis models and adapting them to low-resource languages.

\section{Error Analysis}\label{sec:error_analysis}

In this study, no additional heuristics were applied for tokenizing multiword expressions into separate tokens, which resulted in $19.3$\% of sentences being incorrectly tokenized compared to the ``gold standard''. These sentences were excluded from subsequent error analysis. The analyzed Turkish sentences, from which the annotations were projected onto Kyrgyz, were parsed using the Stanza-IMST model, which demonstrated high quality in the experiment.

Table~\ref{tab:deprel} presents the accuracy of (\texttt{deprel}) tags and the (\texttt{head}) tags.
We consider \texttt{deprel} tags to be correctly predicted if their subtype, when applicable, is also accurately identified. This approach differs from the calculation of LAS and UAS metrics in the UD tools package used in the previous section. The stricter methodology allows for a more detailed assessment of annotation projection quality, which is particularly useful in the context of applying our method to semi-automatic annotation.
The table shows that slightly more than half of the \texttt{deprel} tags were incorrectly predicted in all cases.
It is important to note that the authors of the treebank aimed to showcase the grammatical features of the Kyrgyz language, resulting in the treebank containing a relatively large number of rare and, therefore, challenging cases for automatic syntactic analysis. Moreover, the treebank includes dependency relation types specific to Kyrgyz, such as \texttt{nsubj:pass}, \texttt{nsubj:outer}, \texttt{obl:cau}, \texttt{obl:tmod}, and \texttt{compound:svc}. All these subtypes of syntactic relations (except for \texttt{nsubj:outer}) are not used in the IMST treebank. Consequently, the Turkish model assigned a universal label without specifying a subtype (e.g., \texttt{obl} instead of \texttt{obl:cau}), which was counted as an error during evaluation.

Additionally, a significant percentage of errors stemmed from syntactic differences in sentence translations. Specifically, $35.9$\% of errors in predicting the \texttt{deprel} tag were due to the inability to find a corresponding (i.e. matched in the alignment) word in the Turkish translation for a Kyrgyz word, leaving the Kyrgyz token without a label.

\begin{table}[!t]
    \centering\setstretch{.96}\footnotesize
    \begin{tabular}{lrrr}
    \toprule
        ~ & \textbf{Total} & \textbf{Correct \texttt{deprel}} & \textbf{Correct \texttt{head}}  \\ \midrule
         punct & 152 & 91\% & 57\% \\ 
         nsubj & 118 & 67\% & 64\% \\ 
         root & 117 & 76\% & 76\%  \\ 
         obl & 65 & 58\% & 62\%  \\ 
         aux & 56 & 0\% & 52\% \\ 
         obj & 49 & 67\% & 67\% \\ 
         advmod & 26 & 35\% & 50\% \\ 
         advcl & 23 & 57\% & 52\%  \\ 
         conj & 19 & 53\% & 42\%  \\ 
         nmod & 17 & 18\% & 71\% \\ 
         nmod:poss & 15 & 73\% & 87\% \\ 
         ccomp & 14 & 0\% & 14\% \\ 
         case & 12 & 25\% & 25\% \\ 
         amod & 11 & 73\% & 45\% \\ 
         det & 11 & 55\% & 82\% \\ 
         cc & 10 & 90\% & 40\%  \\ 
         advmod:emph & 10 & 40\% & 50\% \\ 
         acl & 10 & 20\% & 10\%  \\ 
         xcomp & 10 & 0\% & 70\% \\ 
         compound & 10 & 0\% & 10\%  \\ 
         nummod & 7 & 100\% & 100\%  \\ 
         csubj & 7 & 0\% & 43\% \\ 
         fixed & 6 & 0\% & 0\% \\ 
         obl:tmod & 6 & 0\% & 100\%  \\ 
         cop & 6 & 0\% & 83\%  \\ 
         parataxis & 6 & 0\% & 50\% \\ 
         orphan & 5 & 0\% & 40\% \\ 
         flat & 2 & 100\% & 100\% \\ 
         obl:cau & 2 & 0\% & 100\%  \\ 
         mark & 2 & 0\% & 100\%  \\ 
         compound:lvc & 2 & 0\% & 0\%  \\ 
         nsubj:outer & 2 & 0\% & 100\%  \\ 
         discourse & 2 & 0\% & 50\%  \\ 
         compound:svc & 2 & 0\% & 50\%  \\ 
         nsubj:pass & 2 & 0\% & 100\% \\ 
         vocative & 1 & 0\% & 100\%  \\ 
         appos & 1 & 0\% & 0\%  \\ 
         acl:relcl & 1 & 0\% & 100\%  \\
         \bottomrule        
    \end{tabular}
    \caption{Share of the correctly predicted \texttt{deprel} and \texttt{head} labels in the comparison between the \textit{TueCL} annotations and one of the three best models, Stanza$_{\mathrm{IMST,~charlm}}$.}
    \label{tab:deprel}
\end{table}

\begin{figure}[!t]
    \centering
    \resizebox{\linewidth}{!}{
    \begin{tikzpicture}[every node/.style={anchor=base, font=\small}]
    
    \node (s1) at (-2,0) {Питер};
    \node (s2) [right=.005cm of s1] {менен};
    \node (s3) [right=.005cm of s2] {Мэри};
    \node (s4) [right=.005cm of s3] {бири};
    \node (s5) [right=.005cm of s4, red] {-};
    \node (s6) [right=.005cm of s5, red] {бирин};
    \node (s7) [right=.005cm of s6] {кучакташты};
    \node (s8) [right=.005cm of s7] {,};
    \node (s9) [right=.005cm of s8] {андан};
    \node (s10) [right=.005cm of s9, red] {со$\kyrgyzn$};
    \node (s11) [right=.005cm of s10] {б$\kyrgyzo$лм$\kyrgyzo$д$\kyrgyzo$н};
    \node (s12) [right=.005cm of s11] {чыгып};
    \node (s13) [right=.005cm of s12, red] {кетишти};
    \node (s14) [right=.005cm of s13] {.};
    
    \node (t1) at (-1.9,-1) {Peter};
    \node (t2) [right=.34cm of t1] {ve};
    \node (t3) [right=.34cm of t2] {Mary};
    \node (t4) [right=.34cm of t3] {birbirlerini};
    \node (t5) [right=.34cm of t4] {kucaklad{\i}lar};
    \node (t6) [right=.34cm of t5] {ve};
    \node (t7) [right=.34cm of t6] {sonra};
    \node (t8) [right=.34cm of t7] {odadan};
    \node (t9) [right=.34cm of t8] {{\c{c}}{\i}kt{\i}lar};
    \node (t10) [right=.34cm of t9] {.};
    
    \draw[thin] (s1) -- (t1); 
    \draw[thin] (s2) -- (t2); 
    \draw[thin] (s3) -- (t3); 
    \draw[thin] (s4) -- (t4); 
    \draw[thin] (s5) -- (t4); 
    \draw[thin] (s6) -- (t4); 
    \draw[thin] (s7) -- (t5); 
    \draw[thin] (s8) -- (t6); 
    \draw[thin] (s9) -- (t7); 
    \draw[thin] (s10) -- (t7); 
    \draw[thin] (s11) -- (t8);
    \draw[thin] (s12) -- (t9);
    \draw[thin] (s13) -- (t9);
    \draw[thin] (s14) -- (t10);
    
    \draw[->, bend right=90, thin] (s7.north) to (s1.north);
    \node[blue] at ($(s7.north)!0.5!(s1.north)+(0,2.4)$) {nsubj};
    \draw[->, bend right=90, thin] (s3.north) to (s2.north); 
    \node[blue] at ($(s3.north)!0.5!(s2.north)+(0,0.4)$){сс}; 
    \draw[->, bend left=90, thin] (s1.north) to (s3.north); 
    \node[blue] at ($(s1.north)!0.5!(s3.north)+(0,0.95)$){conj};
    \draw[->, bend right=90, thin] (s7.north) to (s4.north);
    \node[blue] at ($(s7.north)!0.5!(s4.north)+(0,1.2)$) {obj};
    \draw[->, bend left=90, thin] (s4.north) to (s5.north); 
    \node[purple] at ($(s4.north)!0.5!(s5.north)+(0,0.2)$){punct}; 
    \draw[->, bend left=90, thin] (s4.north) to (s6.north);
    \node[purple] at ($(s4.north)!0.5!(s6.north)+(0,0.6)$){fixed};
    \draw[->, bend left=90, thin] (s7.north) to (s8.north); 
    \node[purple] at ($(s7.north)!0.5!(s8.north)+(0,0.6)$){punct};
    \draw[->, bend right=90, thin] (s12.north) to (s9.north); 
    \node[blue] at ($(s12.north)!0.5!(s9.north)+(0,1.3)$){obl};
    \draw[->, bend left=90, thin] (s9.north) to (s10.north); 
    \node[purple] at ($(s9.north)!0.5!(s10.north)+(0,0.5)$){case};
    \draw[->, bend right=90, thin] (s12.north) to (s11.north); 
    \node[blue] at ($(s12.north)!0.5!(s11.north)+(0,0.55)$) {obl};
    \draw[->, bend left=90, thin] (s7.north) to (s12.north); 
    \node[blue] at ($(s7.north)!0.5!(s12.north)+(0,2.0)$) {conj};
    \draw[->, bend left=90, thin] (s12.north) to (s13.north); 
    \node[purple] at ($(s12.north)!0.5!(s13.north)+(0,0.6)$) {aux};
    \draw[->, bend left=90, thin] (s7.north) to (s14.north); 
    \node[blue] at ($(s7.north)!0.5!(s14.north)+(0,2.9)$) {punct};
    \draw[->, thin] ($(s7.north)+(0,2.0)$) -- (s7.north);
    \node[blue] at ($(s7.north)+(0,2.2)$) {root};

    \draw[->, bend left=90, thin] ($(t5.north)-(0,0.5)$) to ($(t1.north)-(0,0.5)$);
    \node[blue] at ($(t5.north)!0.5!(t1.north)-(0,2.2)$) {nsubj};
    \draw[->, bend left=90, thin] ($(t3.north)-(0,0.5)$) to ($(t2.north)-(0,0.5)$); 
    \node[blue] at ($(t3.north)!0.5!(t2.north)-(0,0.7)$) {cc}; 
    \draw[->, bend right=90, thin] ($(t1.north)-(0,0.5)$) to ($(t3.north)-(0,0.5)$); 
    \node[blue] at ($(t1.north)!0.5!(t3.north)-(0.2,1)$) {conj};
    \draw[->, bend left=90, thin] ($(t5.north)-(0,0.5)$) to ($(t4.north)-(0,0.8)$);
    \node[blue] at ($(t5.north)!0.5!(t4.north)-(0,1.1)$) {obj};
    \draw[->, bend right=90, thin] ($(t5.north)-(0,0.5)$) to ($(t6.north)-(0,0.5)$);
    \node[blue] at ($(t5.north)!0.5!(t6.north)-(0,0.8)$) {cc};
    \draw[->, bend left=90, thin] ($(t9.north)-(0,0.5)$) to ($(t7.north)-(0,0.5)$); 
    \node[blue] at ($(t9.north)!0.5!(t7.north)-(0,1.3)$) {obl};
    \draw[->, bend right=90, thin] ($(t7.north)-(0,0.5)$) to ($(t8.north)-(0,0.5)$); 
    \node[blue] at ($(t7.north)!0.5!(t8.north)-(0,0.8)$) {case};
    \draw[->, bend right=90, thin] ($(t5.north)-(0,0.5)$) to ($(t9.north)-(0,0.5)$); 
    \node[blue] at ($(t5.north)!0.5!(t9.north)-(0,2.1)$) {conj};
    \draw[->, bend right=90, thin] ($(t5.north)-(0,0.5)$) to ($(t10.north)-(0,0.5)$); 
    \node[blue] at ($(t5.north)!0.5!(t10.north)-(0,2.7)$) {punct};
    \draw[->, thin] ($(t5.south)-(0,2.)$) -- ($(t5.north)-(0,0.5)$);
    \node[blue] at ($(t5.north)-(0,2.9)$) {root};
    \end{tikzpicture}}
    \caption{
    An example from the \textit{TueCL} treebank (at the top) and its Turkish translation (bottom), annotated using the Stanza-IMST-charlm model. Words without a corresponding counterpart in the Turkish sentence are highlighted in red, while dependency relations (\texttt{deprel}) predicted incorrectly are marked in dark red.}\label{fig:tr_ky_errors_example}
\end{figure}
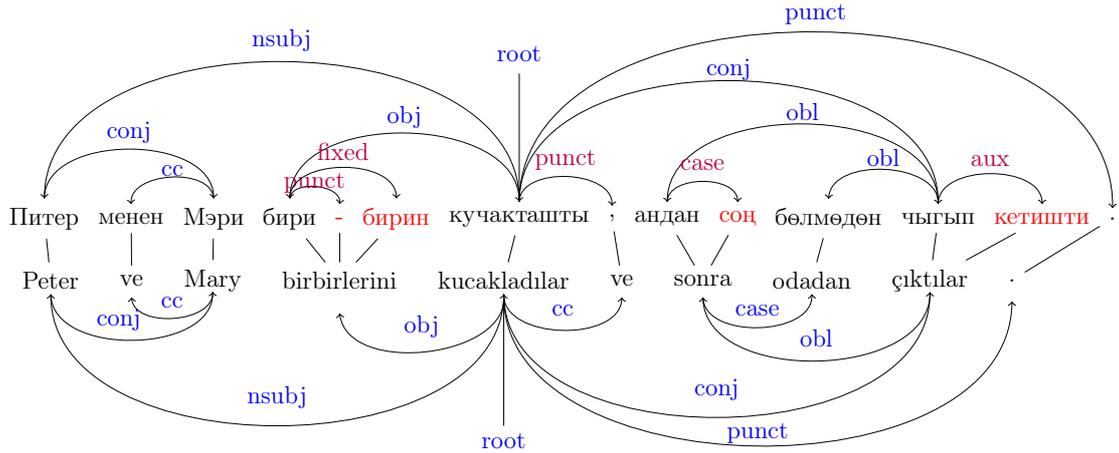

Thus, the \texttt{aux} tag was never predicted correctly (see Table~\ref{tab:deprel}), despite its high frequency in the treebank. In $71$\% of the total errors for this tag, the issue was due to syntactic differences between the languages. Specifically, while the Turkish translation used a synthetic verb form, the Kyrgyz sentence utilized an analytical form, resulting in a discrepancy in the number of tokens. An example of this is shown in Figure~\ref{fig:tr_ky_errors_example}: the reciprocal pronoun ``бири-бирин'' (``each other'') is translated into Turkish as a single word, ``birbirlerini''; ``андан со$\kyrgyzn$'' (``that-ABL after'') is rendered as the single word ``sonra''; and the past tense of the verb is expressed analytically in Kyrgyz as ``чыгып кетишти'' (the synthetic form ``{\c{c}}{\i}kt{\i}lar'' in Turkish).
Additionally, the token for comma is tagged incorrectly, as it is aligned with the conjunction ``ve'' (``and''). For similar reasons, the heads of the clause (\texttt{ccomp}, \texttt{xcomp}, \texttt{csubj}) are often misidentified. In Turkish translations, such constructions are frequently replaced with non-clausal equivalents.

Errors are introduced by the Turkish syntactic parser as well; for instance, all \texttt{parataxis} relations were labeled as \texttt{conj}.

Only $10$\% of errors in determining the \texttt{head} were related to incorrect words alignment. This value is significantly lower compared to the equivalent one for \texttt{deprel} ($35.9$\%) because, for words without a pair during alignment we attached it the root instead of leaving this field empty, as was done for \texttt{deprel}.

\section{Conclusion}\label{sec:conclusion}

\subsection{Key Findings}
This study clearly demonstrates that machine translation combined with dependency parsing models trained on related languages data can significantly expedite manual dependency annotation for the Kyrgyz language. 
Comparison of the approaches to word alignment, syntactic parsing, and machine translation highlights the advantages of employing specialized models in similar tasks. For bitext alignment via multilingual embeddings without any fine-tuning, the XLM-RoBERTa model proved to be the most effective among those considered. For dependency parsing of Turkish texts, the most efficient parser was Stanza-BOUN-BERT. For machine translation, ChatGPT4o with a prompt of task-specific design was the most successful; the prompt instructed the generative model to preserve the word order and the number of words in sentences when translating from Kyrgyz to Turkish.

Given the pressing need to create treebanks for the Kyrgyz language suitable for training parsers, we encourage researchers and practitioners to explore and further develop this approach.

\subsection{Limitations}

When assessing the applicability of the proposed method, the following potential limitations of the presented analysis should be taken into account.

    Firstly, the approach we propose does not incorporate additional language-specific heuristics specifically tailored to adapt annotations projections for the Kyrgyz language. A potential improvement to this tool could involve the use of rules that simplify the annotation process for certain types of words in Kyrgyz, such as discourse markers, copulas, particles, conjunctions, postpositions, etc. Additionally, a specialized method of processing (including the development of tokenization and annotation rules) is required for multiword expressions, which are commonly encountered in the Kyrgyz language.
    
    In the future, the task of multiword expressions tokenization can be addressed using the {apertium-kir} tool~\cite{washington2012finite}, which provides morphological analyses of words within sentences. The development of such heuristics can be carried out on a small dataset of eight sentences\footnote{The annotated sentences are available at \url{https://github.com/ud-turkic/general/blob/main/Annotations/Kyrgyz_JNW.conllu}.}, which were not included in the Kyrgyz language corpus \textit{TueCL} but were also reviewed and annotated as part of the \emph{UD Turkic Group} conference in accordance with the respective treebank guidelines.

    Secondly, this method, despite its simplicity, assumes the availability of a developed morphological analyzer, which is not always available for the ``truly low-resource'' languages.

    The same applies to the use of ChatGPT for sentence translation: for some low-resource languages, the machine translation quality remains low, which can significantly impact the final outcome.

    Thirdly, in this study, we review only a limited number of available models for word alignment and syntactic parsing. Additionally, the question of how the source language influences the generated syntactic annotation remains open. Experiments with other grammatically related languages to Kyrgyz are yet to be conducted in future research.

    There are significant opportunities for the error accumulation at each stage of the proposed processing pipeline. In the future, quality assessment should be performed at \textit{each} step, or at the very least, the impact of each proposed annotation transfer heuristic on the final outcome should be investigated (i.~e. ablation study). However, in this work, the primary focus was on demonstrating the overall feasibility of the proposed approach (\textit{proof-of-concept}). 

    Finally, the \textit{TueCL} treebank is too small for a convincing quality evaluation, its sentences are very similar to each other and have slight variations that highlight the syntactic features of the Kyrgyz language. Therefore, in the future, it would be desirable to use a more representative treebank that includes longer sentences and maintains an equally meticulous annotation. An ideal corpus could collect texts from various genres (registers), such as fiction, popular science, news, encyclopedias, social media, poetry, epic literature, etc.

\subsection{Ideas for Future Research}
The proposed empirical study serves only as a proof-of-concept for the approach, and it is clear that these are the initial steps toward the successful syntactic transfer. In the future, other improvements and alternative strategies can be explored, such as those outlined below.

\emph{Syntax Analysis Using LLMs.} Direct parsing via prompts for large language models (LLMs) represents another promising avenue. Although unpublished results from related research~\cite{tillabaeva2024syntactic} indicate that existing generative models often struggle with zero-shot learning (``prompting'') and few-shot learning (particularly ``prompting with examples'') approaches, we believe that a more in-depth exploration of this method may still uncover its potential for syntactic annotation.

\emph{Automatically Constructed Dependency Trees as a Silver Standard Corpus.} 
The described methods can also be utilized to create a ``silver standard''\footnote{\textit{Silver standard} typically refers to annotation performed by machine learning models with little or partial expert verification.} bank of dependency trees, which could be used for training syntax parsers for the Kyrgyz language. For instance, a Turkish corpus could be translated into Kyrgyz, word alignment (bitext alignment) could be performed, and the annotation could be transferred to the Kyrgyz translation using a method similar to the one proposed. The resulting dataset could then be used to train a parser. While such a parser might not be able to achieve perfect accuracy, it would likely represent a significant improvement in quality compared to the current state of affairs (i.e., the total absence of automatic syntactic parsing or models trained on a small treebank~\cite{benli2023}).

The proposed approaches open promising avenues for addressing the critical shortage of syntactic resources for the Kyrgyz language. By building upon this methodology, future research may further enhance the efficiency and accuracy of the techniques for developing dependency corpora and training parsers, thus advancing both research and practical applications in Kyrgyz language processing as well as other less-resourced languages.

\end{document}